\documentclass[sn-mathphys-num]{sn-jnl}

\usepackage{graphicx}%
\usepackage{multirow}%
\usepackage{amsmath,amssymb,amsfonts}%
\usepackage{amsthm}%
\usepackage{mathrsfs}%
\usepackage[title]{appendix}%
\usepackage{xcolor}%
\usepackage{textcomp}%
\usepackage{manyfoot}%
\usepackage{booktabs}%
\usepackage{algorithm}%
\usepackage{algorithmicx}%
\usepackage{algpseudocode}%
\usepackage{listings}%

\usepackage{array}
\usepackage[caption=false,font=normalsize,labelfont=sf,textfont=sf]{subfig}
\usepackage{textcomp}
\usepackage{stfloats}
\usepackage{url}
\usepackage{verbatim}
\usepackage{arydshln} 
\usepackage{ulem} 
\usepackage{array}
\usepackage{epstopdf}
\usepackage{amsmath, amsfonts}
\usepackage{multirow}
\usepackage{upgreek}
\usepackage{bm}
\usepackage{indentfirst}
\usepackage{color}
\usepackage{amssymb}
\usepackage{arydshln}
\usepackage{adjustbox}
\usepackage{graphicx} 
\usepackage{float} 
\usepackage{booktabs}
\usepackage{tikz}
\usepackage{cite}
\usepackage{pgfplots}
\usepackage{array}
\usepackage{subcaption}
\usepackage{boxedminipage}
\usepackage[numbers]{natbib}
\usepackage{tabularx}

\begin{document}

\title{Progressive Document-level Text Simplification via Large Language Models}








\author[1]{\fnm{Dengzhao} \sur{Fang}}
\author*[1]{\fnm{Jipeng} \sur{Qiang}}\email{jpqiang@yzu.edu.cn}
\author[1]{\fnm{Yi} \sur{Zhu}}
\author[1]{\fnm{Yunhao} \sur{Yuan}}
\author[2]{\fnm{Wei} \sur{Li}}
\author[1]{\fnm{Yan} \sur{Liu}}

\affil*[1]{\orgdiv{School of Information Engineering}, \orgname{Yangzhou University}, \orgaddress{\street{}, \city{Yangzhou}, \postcode{225127}, \state{Jiangsu}, \country{China}}}

\affil[2]{\orgdiv{School of Artificial Intelligence and Computer Science}, \orgname{Jiangnan University}, \orgaddress{\street{}, \city{Wuxi}, \postcode{214122}, \state{Jiangsu}, \country{China}}}


\abstract{Research on text simplification has primarily focused on lexical and sentence-level changes. Long document-level simplification (DS) is still relatively unexplored. Large Language Models (LLMs), like ChatGPT, have excelled in many natural language processing tasks. However, their performance on DS tasks is unsatisfactory, as they often treat DS as merely document summarization. For the DS task, the generated long sequences not only must maintain consistency with the original document throughout, but complete moderate simplification operations encompassing discourses, sentences, and word-level simplifications. Human editors employ a hierarchical complexity simplification strategy to simplify documents. This study delves into simulating this strategy through the utilization of a multi-stage collaboration using LLMs. We propose a progressive simplification method (ProgDS) by hierarchically decomposing the task, including the discourse-level, topic-level, and lexical-level simplification. Experimental results demonstrate that ProgDS significantly outperforms existing smaller models or direct prompting with LLMs, advancing the state-of-the-art in the document simplification task.}

\keywords{Document Simplification, Large Language Model, ChatGPT, Multi-stage System, Text Generation}



\maketitle

\section{Introduction}
\begin{sloppypar}
Text simplification aims to simplify the input text by reducing its complexity to make it more understandable for a wider audience, including non-native speakers \citep{paetzold2016unsupervised} and individuals with cognitive impairments \citep{ gooding-2022-ethical,kajiwara2013selecting, paetzold2015reliable}.  Lexical-level and sentence-level simplification have been the main focus in the field of text simplification by training neural network models or fine-tuning pre-trained language models using supervised data \citep{paetzold2016unsupervised,gooding-2022-ethical}. Since the challenge of generating long sequential text output has persisted, document-level simplification (DS) has attracted little attention. Traditional approaches for document simplification often rely on simplifying sentences individually, without considering the broader context of the document. This disregard for the overall document context can lead to a loss of coherence and integrity in the simplified version.

Recently, these DS work \citep{cripwell2023document, Sun2023TeachingTP} primarily focused on sentence-level simplification operations (copy, rephrase, split, or delete). Binova et al. \citep{Blinova2023SIMSUMDT} proposed a two-stage framework, which first generated one summary of the input document and generated the simplified summary. The recent development of Large Language Models (LLMs), such as ChatGPT, has ushered in a new paradigm in natural language processing \citep{zhou2023comprehensive}. 
Several studies \citep{Feng2023SentenceSV,kew-etal-2023-bless} have highlighted the significant potential of Large Language Models (LLMs) in sentence simplification, requiring no specialized training data or modifications to model parameters.  

\begin{table}[h!]
\centering
\caption{\label{tab:len_sta}
Comparison of token counts between manually generated simplified documents and those generated by ChatGPT on Newsela. Newsela-A and Newsela-B are composed of 500 sample pairs from the Newsela dataset with article lengths below 1000 tokens and with article lengths exceeding 1,000 tokens. For each original article (SRC),  three simplified versions (RE1, RE2, and RE3) by humans are shown here, where RE3 is the simplest. 'P1', 'P2', and 'IC'  represent three simplified versions generated by ChatGPT guided by three different prompt templates. }

\begin{tabular}{l|c|c|c}\hline
 \textbf{Dataset} & \textbf{SRC} & \textbf{RE1(RE2, RE3)} & \textbf{P1(P2,IC)}\\\hline
 Newsela-A & 818 & 833 (805, 768) & 352(334, 308) \\
 Newsela-B & 1506 & 1286 (995, 863) & 445(329, 378) \\
\hline
\end{tabular}

\end{table}

We test ChatGPT on one DS data Newsela \citep{xu2015problems} constructed by humans, where each article of Newsela was rewritten four times by professional editors for children at different grade levels. As shown in Table \ref{tab:len_sta}, ChatGPT with three different prompts commonly produces a summary of the original document rather than a simplification of the document. The most obvious sign is that the length of the simplified output is much shorter than the references. Although we tried three different prompting strategies including direct prompting (P1), pointing out the difference between summarizing and simplifying (P2), and few-shot setting (IC), they still yielded unsatisfactory results. Why do LLMs work well for document summarization and not for DS tasks? DS has the following two challenges when applied to LLMs. 

(1) Content Preservation:
While simplifying text, it is important to retain the essential information and meaning of the original document. But, while generating text in lengthy sequences, it is a formidable challenge to decide which details to preserve and which to simplify or omit to maintain the document's overall integrity and message.

(2) Ambiguity and Subjectivity: Simplification involves making decisions about how to convey information in a clearer and more straightforward manner. Large language models may struggle to consistently produce simplifications that meet human expectations and preferences, especially for longer texts where ambiguity and nuance are more prevalent.

We see that LLMs fall short of achieving DS with a single prompt. LLM-based multi-stage systems have achieved considerable progress in complex problem-solving \citep{guo2024large}. This paper aims to explore the design of a multi-stage-based approach that can imitate human editors' approach to simplifying long documents using LLMs. The human editor accomplishes the simplification of the document through the following steps: (1) Before simplifying the text, it's crucial to understand the document's purpose and identify the main topics or sections of the document. (2) Then, the content of each topic is simplified individually, including sentence rephrasing, deletion, lexical simplification, and other operations. Therefore, to simulate this simplified process of human editors, we propose a progressive simplification method (ProgDS) that enables LLMs to follow human-simplified operations under multi-stage collaboration.

Specifically, ProgDS dissects the task of DS into three hierarchical steps: discourse-level, topic-level, and lexical-level simplification. Starting from the overall logical structure, then moving on to the arrangement and combination of paragraphs and sentences, and finally to lexical expressions, this task is approached gradually and systematically, simplifying each level step by step, rather than attempting to simplify all these levels simultaneously. 

In summary, the contributions of this paper are:

(1) To our knowledge, we are the first to solve the task of long document-level simplification through the utilization of a multi-stage collaboration using LLMs. This bridges the gap between text simplification and practical applications with the assistance of LLMs. 

(2) We propose one novel DS method ProgDS by executing hierarchical complexity simplification. This alleviates the limitations of LLMs unable to simplify long documents through direct prompting. Our frameworks follow the simplification methods of human experts, integrating discourse-level, topic-level, and lexical-level simplification based on the principles of content and hierarchical division. 

(3) We evaluate ProgDS compared with existing methods on Wiki-auto and Newsela datasets. Experimental results show that ProgDS achieves state-of-the-art performance across various evaluation metrics. The benefits of our algorithm become more pronounced, especially when dealing with longer original documents.
\end{sloppypar}

\section{Related Work}

\subsection{Document-Level Text Simplification} 

The field of text simplification has traditionally focused on lexical and sentence-level methods \citep{xu2015problems,gooding-2022-ethical}. While sequence-to-sequence frameworks are common in sentence simplification, they struggle with document simplification due to limited supervised data. Document simplification requires holistic consideration of semantic content, involving operations like sentence simplification, deletion, retention, and merging. Efforts to extract training corpora from sources like English Wikipedia (EW) and Simple English Wikipedia (SEW) face challenges due to suboptimal corpus quality.

Recent advancements include the SWIPE dataset \citep{laban-etal-2023-swipe}, which leverages complete revision histories to improve page pairing and identify effective edits. This dataset primarily enhances sentence-level editing operations. Meanwhile, Sun et al.~\citep{Sun2023TeachingTP} proposed a continuous pre-training strategy for SimpleBART, focusing on sentence-level simplification rather than long-text documents.

In document-level simplification, Cripwell et al.~\citep{cripwell2023document} predicted sentence edits based on document context, simplifying each sentence individually. In contrast, Blinova et al.~\citep{Blinova2023SIMSUMDT} emphasized summarizing the document first and then simplifying the summary, deviating from traditional document simplification. Inspired by these approaches, a new method integrates summarization to guide document simplification, preserving the document's core topic while ensuring overall simplification.

\subsection{Text Simplification using LLMs} 

Currently, there has been a notable increase in the development of Large Language Models (LLMs) through extensive training on vast quantities of textual data. These LLMs including GPT-3 \citep{2020Language} and ChatGPT have showcased exceptional capabilities in generalizing information and have achieved impressive results across a range of specific applications. 

Several studies have investigated the use of large language models (LLMs) in text simplification. Feng et al.~\citep{Feng2023SentenceSV} compared ChatGPT with traditional fine-tuning methods for sentence simplification and found ChatGPT to be superior in both automated and human evaluations, outperforming most pre-training and fine-tuning approaches. However, Sun et al.~\citep{Sun2023TeachingTP} and Laban et al.~\citep{laban2023swipe} assessed ChatGPT's performance in paragraph-level text simplification and discovered that it did not surpass traditional fine-tuning methods according to automated metrics. They speculated that in zero-shot settings, ChatGPT struggles with longer texts due to a lack of specific knowledge about the simplification format and degree, leading to discrepancies between generated and reference simplifications.

Moreover, Raheja et al.~\citep{raheja2023coedit} and Shu et al.~\citep{shu2023rewritelm} performed instruction-tuning on several open-source LLMs, enhancing the models' ability to follow user text editing instructions, including those for text simplification. These models showed promising results across various text editing tasks. However, these instruction-based fine-tuning methods necessitate significant amounts of data, detailed instructions, and substantial computational resources.

\subsection{Hierarchical Text Generation}

Long text generation poses challenges for current language models in maintaining coherence and context as the text becomes longer\citep{drissi2018hierarchical}. While recurrent models theoretically have the capacity to handle long sequences, they often struggle in practice. Attention-based models like the Transformer \citep{vaswani2017attention} can perform parallel computations on the entire input sequence, but they are typically limited by computational resources and cannot handle excessively long sequences. Therefore, existing language models often struggle to read and generate longer sequences, or they typically process text sentence by sentence, resulting in strong local coherence but weak global coherence. Hierarchical text generation is proposed as a solution to address these difficulties, aiming to improve the model's ability to keep track of context and generate text that is both locally and globally coherent.

Recently, Zhou et al.~\citep{zhou2023recurrentgpt} based on LLMs generates long text by combining input paragraphs, plans, and memories, and uses LLMs to generate new paragraphs while updating its memories. It allows for generating text of any length and enables human interaction and manipulation of memories and plans. Puduppully et al.~\citep{puduppully2019datatotext} attempted to use similar methods in data-to-text generation tasks. They first generated an overall content plan based on the data, and then generated specific textual descriptions according to the content plan. They ultimately demonstrated that this approach helps improve the conciseness and coherence of generated content while reducing hallucinations. In this paper, we focus on document-level simplification based on LLMs.

\section{Methods}

In this paper, document-level text simplification (DS) is formulated as a conditional generation problem, where a language model generates a simplified version \textbf{Y} autoregressively conditioning on the input source document \textbf{X}. Document summarization and DS are related but distinct tasks in natural language processing. Document summarization condenses the document into a shorter version while retaining essential content, whereas DS aims to simplify the entire document, making it more accessible and easier to understand without necessarily reducing its length significantly. 

Given the input source document $\textbf{X}$ and an autoregressive modeling $g$ with parameters $\theta$, the model will output the simplified document $\textbf{Y}=\left\{y_1, y_2, \cdots, y_m\right\}$ with $m$ tokens. This process can be formalized as:

\begin{equation}
p_g(\mathbf{Y} \mid \mathbf{X}, \theta)=\prod_{t=1}^m p_g\left(y_t \mid \mathbf{Y}_{<t}, \mathbf{X}, \theta\right),
\end{equation}
where $y_t$ represents the current generated token by considering the previously generated context $\mathbf{Y}_{<t}$, and $p_g$ is the probability distribution parameterized by the modeling $g$.

\subsection{One Simple Method: SumDS}

\begin{figure}[t!]   
    \centering
    \includegraphics[width=1.0\linewidth]
    {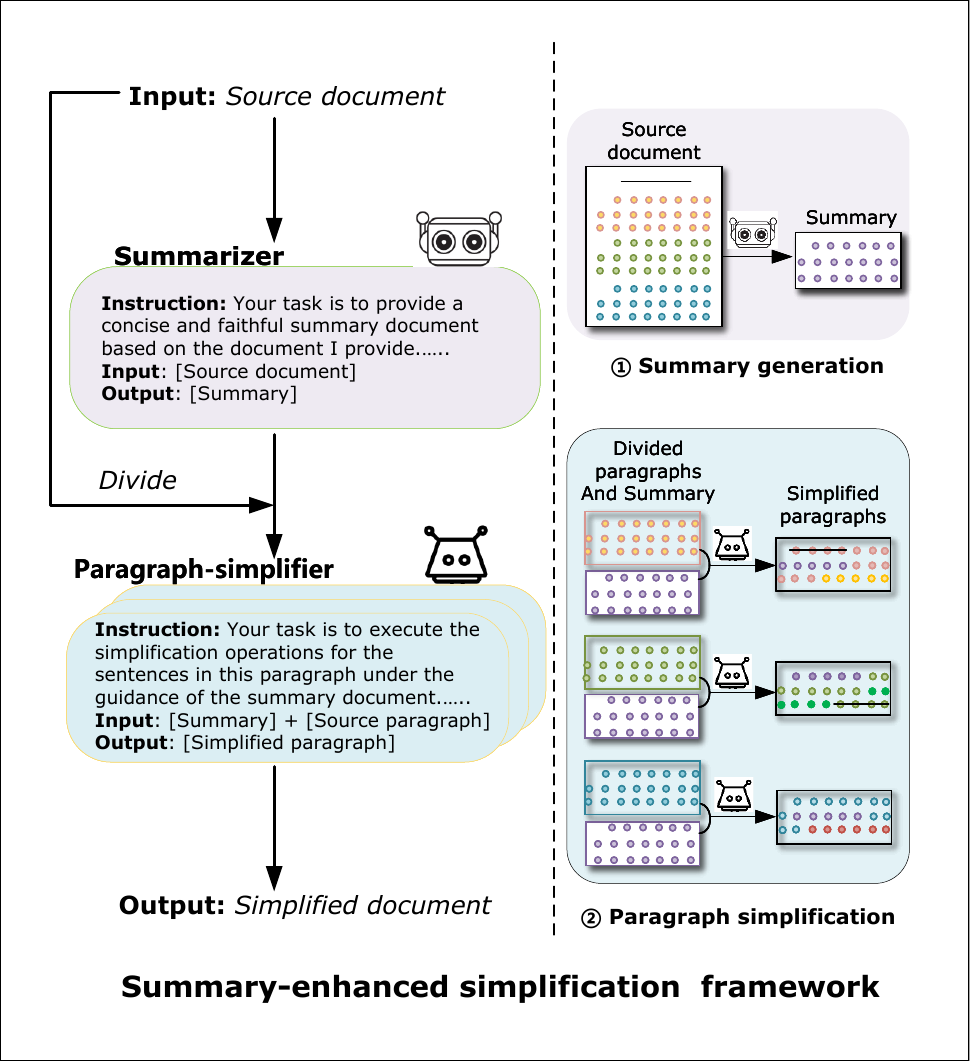}
    \caption{The framework of Summary-enhanced simplification(SumDS) to generate the simpler version. After dividing the source document based on its content, it is simplified separately guided by the summary. The simplified segments are then concatenated to form the final output.}
    \label{fig:SumDS}
\end{figure}

 The overall framework of SumDS is shown in Figure \ref{fig:SumDS}. SumDS first generates one summary for the source document, and performs simplification operations for each paragraph under the guidance of the summary. During the simplification process of each paragraph, all the sentences (especially during sentence deletion and preservation operations) not only consider the contextual information from the summary document but also take into account the information from surrounding sentences within the paragraph. 

\begin{figure}[t!]
\centering
\begin{boxedminipage}{\columnwidth}
\footnotesize
Please generate a summary of this document using the shortest possible language, retaining only the most important information.\\
Source document: {\color{blue}{\text{[Source document]}}}\\
Summary:
\end{boxedminipage}
\caption{The prompting template of \textit{Summarizer} of SumDS, where the contents within "[]" are variables.}
\label{fig:prompt_sum}
\end{figure}

\textbf{Summary Generation.} Given the outstanding performance of LLMs in automatic text summarization, we draw inspiration from Zhang et al.~\citep{zhang2023benchmarking}'s prompt template to guide ChatGPT in generating the summary, shown in Figure \ref{fig:prompt_sum}.

\begin{figure}[t!]
\centering
\begin{boxedminipage}{\columnwidth}
\footnotesize
You are a text editor tasked to simplify a document. You goal is to simplify paragraphs under the guidance of the summary. Here are some operations that may be used:\\
    1. Delete irrelevant sentences based on the summary document and context.\\
    2. Merge complex and redundant sentences to improve readability.\\
    3. Split complex sentences into simpler ones.\\
    4. Rephrase sentences with complex words or phrases.\\
    5. Retain important and already simplified sentences.\\
    6. Replace difficult expressions with simpler ones.\\
    {\color{blue}{\text{[Examples]}}}\\
    Summary: {\color{blue}{\text{[The generated summary]}}}\\
    Paragraph to be simplified: {\color{blue}{\text{[Paragraph to be simplified]}}}\\
    Simplified paragraph:
\end{boxedminipage}
\caption{The prompting template of \textit{Paragraph-simplifier} of \textit{SumDS}. "(Examples)" refers to the examples needed for few-shot learning and chain-of-thought.}
\label{fig:prompt_para}
\end{figure}

\textbf{Paragraph Simplification.}  We divide the source document into multiple paragraphs and proceed to simplify them one by one. The prompt template of paragraph simplification is shown in Figure \ref{fig:prompt_para}. Guided by the summary and the source paragraph, the prompting involves compressing multiple sentences into one, splitting a single sentence into multiple sentences, deleting or retaining certain sentences, and using simpler expressions to replace complex ones, among other simplification operations. 

It is worth noting that in the case of a document with only one paragraph to be simplified, we divide it into multiple sentences and perform the simplification under the supervision of the summary, using a prompt template similar to the version mentioned in Figure \ref{fig:prompt_para}.

\subsection{Progressive Simplification: ProgDS}

\begin{figure}[t!]   
    \centering
    \includegraphics[width=\linewidth]{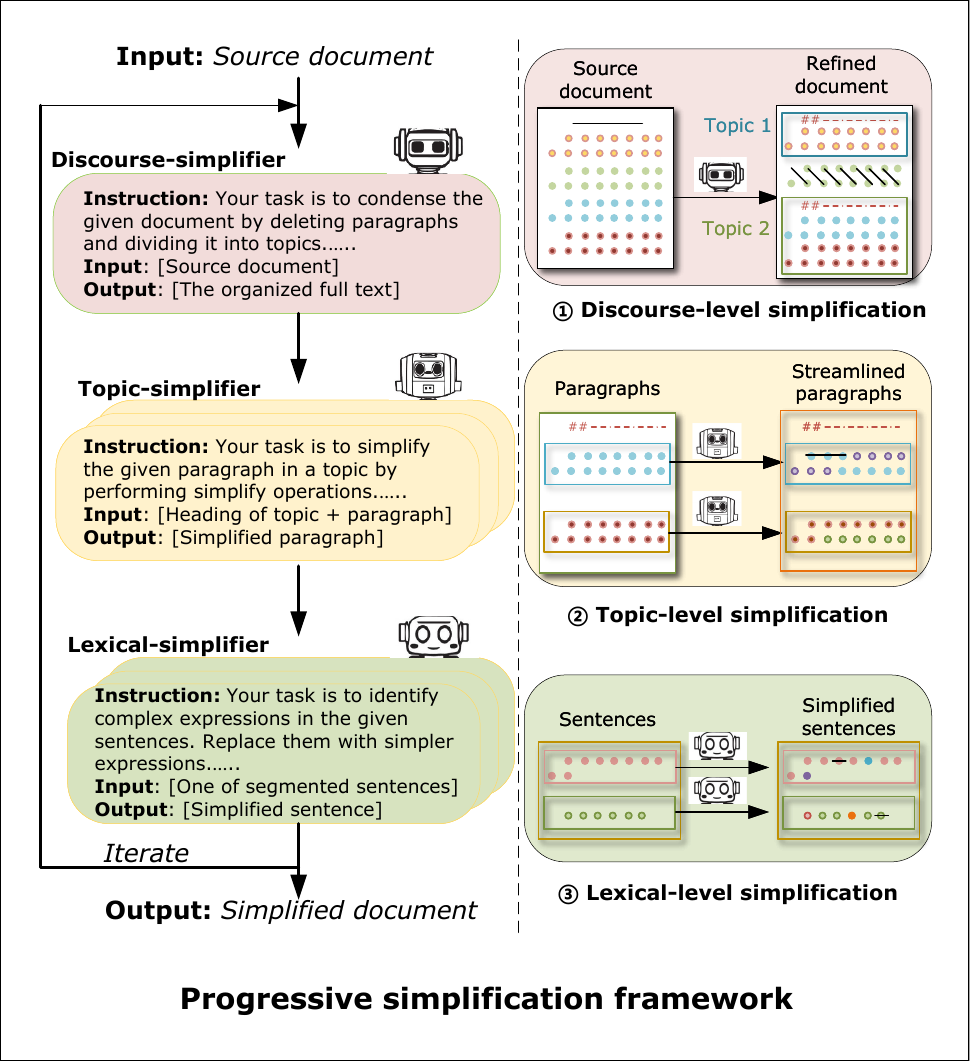}
    \caption{The framework of progressive simplification (ProgDS). The three levels of discourse-level, topic-level, and lexical-level simplification are performed sequentially. Moreover, the topic-level and lexical-level simplification are executed multiple times within a document.}
    \label{fig:ProgDS}
\end{figure}

SumDS based on contextual guidance has an inherent limitation: it rigidly concatenates individually simplified paragraphs, which may result in the loss of coherence in the original text. In addition, there is also the problem of incomplete simplification. Considering the aforementioned limitations, we attempted to design a framework based on LLMs that is more suitable for simplifying long documents. We have considered the practices of human experts when performing long document-level text simplification. Experts often follow a hierarchical approach to text editing: Starting from the overall logical structure, then moving on to the arrangement and combination of paragraphs and sentences, and finally to lexical expressions. This task is approached gradually and systematically, simplifying each level step by step, rather than attempting to simplify all these levels simultaneously. Therefore, we propose one progressive simplification method (ProgDS) by decomposing the task and executing it in a hierarchical manner, including the discourse-level, topic-level, and lexical-level simplification.

Furthermore, the simplified documents written by human experts are divided into multiple levels of simplification. To achieve higher levels of simplification, we also use an iterative approach to further simplify the previous version. The overall framework of this method is shown in Figure \ref{fig:ProgDS}.

\begin{figure}[t!]
\centering
\begin{boxedminipage}{\columnwidth}
\footnotesize
 You are a professional manuscript editor and reviewer. The task is to organize and divide an article into multiple distinct topics. Each paragraph in the article is numbered. \\
  1. The goal is to maintain a consistent central theme for each topic. \\
  2. Subheadings need to be generated for each topic.\\
  3. Irrelevant paragraphs can be deleted.\\
  The output format must be a subheading followed by paragraph numbers, where these paragraph numbers represent the same topic.\\
    {\color{blue}{\text{[Examples]}}}\\
    Source document: \\
    {\color{blue}{\text{[Source document with paragraph number]}}}\\
    The organized content:
\end{boxedminipage}
\caption{The prompting template of \textit{Discourse-simplifier} in ProgDS.}
\label{fig:discourse_level}
\end{figure}

\textbf{Discourse-level simplification.} According to Mathias et al.~\citep{mathias2018eyes}'s assessment of the quality of article readability, three different aspects were considered: organization, cohesion, and coherence. We are committed to enhancing the quality of logical structure in articles in these three aspects as much as possible. Regarding these three assessment aspects, they point out that texts that are poorly organized or lack cohesion can make readers go back to previous sentences or paragraphs. When a text lacks coherence, readers may focus more on different parts of the text in order to understand them. Therefore, a popular and easy-to-understand article must first have a well-structured and coherent organization throughout its discourse.

In this stage, we use LLM to divide the entire document into multiple topics, where each topic consists of one or multiple paragraphs. This process contains two tasks: text clustering and subheading generation. For each topic, it is also necessary to generate a corresponding subheading placed above the topic. The placement of these subheadings can make the overall structure of the article clearer and more visually intuitive, enabling readers to grasp the overall context of the article quickly.

Considering the possibility of input and output containing an excessive number of tokens, we cannot directly output the words when designing prompt templates for discourse-level simplification. We first involve labeling each paragraph in the source document with sequential numbers, and then instruct LLM to correspond the subheadings with the given numbers when generating the output. Furthermore, for redundant paragraphs, LLM can decide to delete them. In fact, human experts often delete entire paragraphs when performing document-level simplification. The prompting template is shown in Figure~\ref{fig:discourse_level}. For source documents consisting of only one paragraph, we assign a number to each sentence and group multiple sentences into one topic, and use another similar prompt template.

\begin{figure}[t!]
\centering
\begin{boxedminipage}{\columnwidth}
\footnotesize

You are a professional manuscript editor and reviser. The task is to simplify the given paragraph under the guidance of the subheading associated with the topic of the paragraph to enhance accessibility and readability.\\ 
When it comes to the meaning and structure of the entire paragraph, you need to follow these tips:\\
  1. Identify key points and simplify the structure.\\
  2. Offer extra context for unfamiliar concepts.\\
  3. Maintain logical flow and consider paragraph division.\\
  {\color{blue}{\text{[Examples]}}}\\
  When it comes to the structure between sentences and within individual sentences, you need to follow these tips:\\
  1. Combine simple sentences.\\
  2. Divide complex sentences into simpler ones.\\
  3. Remove irrelevant sentences.\\
  4. Rearrange sentence order for better flow.\\
  5. Utilize basic subject-verb-object sentence structure.\\
    {\color{blue}{\text{[Examples]}}}\\
    Subheading of current topic:{\color{blue}{\text{[Subheading of current topic]}}}\\
    Paragraph to be simplified: {\color{blue}{\text{[Paragraph to be simplified]}}}\\
    The simplified paragraph:
\end{boxedminipage}
\caption{The prompting template of \textit{Topic-simplifier} in ProgDS.}
\label{fig:para_level}
\end{figure}

\textbf{Topic-level simplification.} After the discourse-level simplification in the previous step, the article has become reasonably detailed and well-structured as a whole, each article contains multiple topics, with each topic corresponding to a subheading and several paragraphs. Next, we formulate prompt templates to guide the model in simplifying each paragraph under a given topic with the instruction of a simple subheading. For source documents consisting of only one paragraph, in the previous step, it has been divided into multiple topics with each topic containing several sentences, we consider multiple sentences under each topic as a paragraph. 

Specific operations regarding paragraph and sentence structures need to be completed simultaneously in this stage. Regarding the simplification of paragraph content and structure, some ideas and actions include the following aspects: determine the key points or concepts conveyed in the whole paragraph, focus on preserving these important elements while simplifying the structure; if one paragraph refers to specific concepts, events, or ideas that may be unfamiliar to the reader, provide additional context or explanations to enhance understanding; ensure that sentences within a paragraph are logically connected, use appropriate transition words or phrases to guide the reader through the flow of ideas; if a paragraph contains multiple ideas or information, consider dividing it into smaller paragraphs.

Regarding the simplification of the structure between sentences and within individual sentences include the following aspects: multiple simple and repetitive sentences can be concatenated into one sentence, while a complex and lengthy sentence can be split into multiple simple sentences; sentences that have little relevance or impact on the context can be deleted, while important and structurally simple sentences can remain unchanged; convert complex or compound sentences into simpler ones, instead of using multiple clauses or phrases, opt for shorter, straightforward sentences, the subject-verb-object structure (e.g., ``\textit{The cat chased the mouse.}”) is often easier to understand. The prompting template is as shown in Figure~\ref{fig:para_level}.

\begin{figure}[t!]
\centering
\begin{boxedminipage}{\columnwidth}
\footnotesize
You are a query engine equipped with a wide range of simpler alternatives for complex expressions. The task is to identify complex and uncommon vocabulary, phrases, idioms, etc. in a given sentence. And then provide simplified alternatives for these complex elements.\\
  1. Incorporate the replacements into the sentence and ensure that the sentences remain smooth and coherent.\\
  2. Explain an unfamiliar idea using more familiar words and examples that people know.\\
  3. The sentence structure doesn't need to be considered, and the overall meaning should be maintained as much as possible after the replacements are made.\\
    {\color{blue}{\text{[Examples]}}}\\
    Sentence to be simplified:\\
    {\color{blue}{\text{[Sentence to be simplified]}}}\\
    The simplified sentence:
\end{boxedminipage}
\caption{The prompting template of Lexical-simplifier used in ProgDS.}
\label{fig:lexical_level}
\end{figure}

\textbf{Lexical-level simplification.} After the two stages mentioned above, the last stage is the lexical simplification. In this stage, the logical structure of the article, paragraph, or sentence is no longer considered. The focus is solely on the complexity of the expressions such as vocabulary, phrases, and idioms. Simplifying vocabulary is necessary for targeting specific groups like children or non-native speakers who may struggle with complex vocabulary or idiomatic expressions because of their limited vocabulary. The importance of lexical simplification is even higher than the previous two stages because relevant research in psycholinguistics has found that as long as the readers are familiar with the vocabulary, they can still understand the meaning of the article even if they cannot understand some of the grammar in the article \citep{nation2001learning}. 

Lexical simplification typically involves three steps \citep{qiang2020lexical}: identifying complex words, generating the candidate substitutes, and selecting the most appropriate substitute word. Within the framework based on LLMs, these operations can be performed simultaneously by simply writing appropriate prompt templates, with the core focus still being on substitution, replacing complex expressions with simple and commonly used ones can significantly enhance the readability of the article. 

Instead of using uncommon or idiomatic expressions, it is better to use common phrases and idioms. Opt for more common and straightforward phrases. For example, replace "\textit{utilize}" with "\textit{use}", and avoid idioms like "\textit{piece of cake}" by using simpler language such as "\textit{easy}". In addition, there are also some advanced ways of expressing emotions, such as irony and so on. These are expressions that beginner language learners find difficult to comprehend and need to be replaced with more straightforward and direct expressions. We guide the model in recognizing and replacing these complex expressions based on specific instruction patterns. The prompting template is shown in Figure~\ref{fig:lexical_level}.

\subsection{In-Context Learning}\label{sec:ICL}

\begin{figure}[t!]
\centering
\begin{boxedminipage}{\columnwidth}
\footnotesize
  To transform a complex and difficult-to-understand sentence into a simple and easy-to-understand one requires a certain thought process. Now, please learn from some examples and provide the thought process for the given complex-simple sentence pairs.\\
    Complex sentence: A story of animals healing depression.\\
    Reasoning: The phrase ‘healing depression’ is a bit difficult to understand, depression, as mentioned here, can be described as a condition that brings about feelings of sadness, and animals can help in treating this condition.\\
    Simple: A story about animals that can help when someone feels sad.\\
    {\color{blue}{\text{[Complex-simple sentence pair]}}}\\
    The reasoning of this pair:
\end{boxedminipage}
\caption{Example of using prompting template to generate chain-of-thought for existing complex-simple sentence pairs.}
\label{fig:COT}
\end{figure}

In-context learning (ICL) has gained popularity due to its effectiveness and efficiency in leveraging LLMs. ICL involves the selection of informative demonstrations, which are used as additional input to improve the performance of LLMs. This technique aims to enhance the results by leveraging semantically similar examples (few-shot) or utilizing uncertainty and diversity for demonstration refinement and evaluation. The chain-of-thought (COT) technique \citep{fu2023complexitybased} incorporates reasoning complexity into the demonstration process enabling a more comprehensive understanding of the reasoning process.

When designing our framework, we write or select relevant examples to enable LLMs for few-shot learning. We also find that guiding the model through COT technique can enhance the quality of text revisions. It is crucial to provide comprehensive and clear instructions to LLMs, we draw inspiration from the requirements given to human writers on Simple English Wikipedia~\footnote{\url{https://simple.wikipedia.org/wiki/Wikipedia:How\_to\_write\_Simple\_English\_pages}}.

For specific instruction descriptions, we input them into the \textit{system} role of the interface of GPT-3.5~\footnote{\url{https://platform.openai.com/docs/api-reference/chat/create}}). For example, "\textit{You are a professional manuscript editor and you are required to simplify the given article...}" (Omitted complete content). As for examples of few-shot learning and chain-of-thought, we input them into the \textit{user} role of the interface. For example, when performing lexical simplification, the "(examples)" in Figure~\ref{fig:lexical_level} can be "\textit{Complex sentence: A story of animals healing depression. Reasoning: The phrase ‘healing depression’ is a bit difficult to understand...... Simple sentence: A story about animals that can help when someone feels sad.}" . In this example, the "Complex sentence" and "Simple sentence" represent the examples for few-shot learning, while "Reasoning" represents the example for COT.

When it comes to the summary generation stage in SumDS, we use zero-shot learning, because LLMs have demonstrated sufficiently good zero-shot learning performance on this task. For the paragraph simplification stage in SumDS, we select examples based on vector similarity~\footnote{We use the \textit{paraphrase-MiniLM-L6-v2} model to encode sentences and implement it using the \textit{sentence\_transformers} library in Python.} from Wiki-auto\citep{jiang-etal-2020-neural} because most samples in Wiki-auto are relatively simple and suitable for paragraph-level simplification.

For the discourse-level simplification in ProgDS, We use manually crafted examples to instruct LLMs on the tasks they need to accomplish at this stage and the desired output format they should obtain. For the topic-level simplification in ProgDS, we divide it into two aspects of simplification and provide ICL prompts. Firstly, for the aspect of paragraph meaning and structure, we use a vector similarity method to extract samples from wiki-auto as examples. Secondly, for the aspect of internal sentence structure, we use a structure similarity method to extract samples from ASSET\citep{alva-manchego-etal-2020-asset} as examples. Here, we calculate sentence structure similarity using factors including length, clause numbers, parts of speech, and structure, taking the average value of sentences as the value of the paragraph.  The samples with higher similarity are more suitable to be used as examples. For lexical-level simplification in ProgDS, we also filter suitable examples from the LexMTurk\citep{horn2014learning} and BenchLS\citep{paetzold2016unsupervised}, the approach primarily involves providing more simplified transformations of different part-of-speech in the ICL prompts.

The examples given above do not show the reasoning process behind them, so they only meet the need for few-shot learning examples, not the need for COT examples. Manually writing out the thought process for these existing examples can be time-consuming. So instead, we use LLMs to automatically generate them. We prime the LLMs with well-written examples that show the thought process. As shown in Figure~\ref{fig:COT}, this helps the LLMs learn from these examples to generate the thought process behind the existing examples.

It is worth noting that although we have decomposed the tasks and provided detailed and comprehensive task descriptions and in-context learning examples, the model may still generate outputs that do not strictly adhere to the format or requirements. To solve this problem, we employ an \textit{over-generate-then-filter}\citep{wiegreffe-etal-2022-reframing} method to select the desired outputs that meet the requirements. LLMs often yield different outputs for the same input. The main idea is to specify a rule~\footnote{We use regular expressions implemented in Python to filter the output of the model, primarily including restricting output formats and extracting the required results from the model's output.} to filter out outputs that do not meet the required format. Once we obtain the desired output, we stop further generation. 

\section{Experiments}

\subsection{Experiment Settings}

\begin{table}[t!]
\centering

\begin{tabular}{c|cccc}
\toprule[1pt]
     & Wiki-auto & Newsela-A & Newsela-B \\
    \midrule
    Paragraphs-X & 1 & 17.16 & 28.94  \\
    Paragraphs-Y & 1 & 16.41 & 20.45  \\
    \midrule
    Sentences-X & 11.62 & 30.60 & 55.31  \\
    Sentences-Y & 9.20 & 43.80 & 51.69  \\
    \midrule
    Tokens-X & 378.62 & 818.33 & 1506.70 \\
    Tokens-Y & 230.75 & 769.79 & 976.36 \\
    
\bottomrule[1pt]
\end{tabular}

\caption{\label{tab:data}  Detailed information statistics for the three divided and extracted datasets, where X represents the source document, Y represents the reference simplified document (the statistical information for the reference document in the Newsela dataset is the average of multiple samples).}
\end{table}

\textbf{Dataset.} We mainly utilize the Newsela-auto \citep{jiang-etal-2020-neural}, which is currently the most suitable dataset for document-level text simplification, and one dataset extracted from Wikipedia Wiki-auto \citep{jiang-etal-2020-neural}. Additionally, we refer to Sun et al.~\citep{sun2023pearl}'s definition of long documents and selected articles with over 1,000 tokens from the Newsela-auto dataset as our main test dataset.

The first dataset consists of 500 randomly selected sample pairs from Wiki-auto with text lengths ranging from 300 tokens to 500 tokens. Newsela-A and Newsela-B are composed of 500 sample pairs from the Newsela dataset with article lengths below 1000 tokens and with article lengths exceeding 1,000 tokens. The statistical information of these datasets can be found in Table \ref{tab:data}.

\textbf{Metrics.} Based on factors such as simplicity, completeness, fluency, and overall score, we select a total of five evaluation metrics, including four computational metrics and one AI self-assessment metric.

(1) SARI based on $n$-gram edit calculation is commonly used metric for sentence-level \citep{Xu2015ProblemsIC}. (2) D-SARI is a modified indicator based on SARI that penalizes the three components in SARI, specifically suitable for simplified evaluation at the document level \citep{Sun2021DocumentLevelTS}. (3) BARTScore(BART-S) is employed to evaluate the preservation of meaning and fluency in the generated text \citep{yuan2021bartscore}. (4) Flesch-Kincaid grade level (FKGL) is a commonly used indicator for assessing the readability of texts\citep{scialom2021rethinking}.

(5) GPT-Evaluate(GPT-E) enables ChatGPT to compare and score the generated text in pairs. For simplified texts generated using different methods, we combine them with directly prompted simplified texts obtained from ChatGPT to form prompt templates and have ChatGPT compare and score them, taking the average win rate as the result. The prompting template of GPT-Evaluate is as shown in Figure \ref{fig:Judge}. 

\begin{figure}[t!]
\centering
\begin{boxedminipage}{\columnwidth}
\footnotesize
  You are a professional document review expert with a strong foundation in writing and extensive experience in reviewing. Please compare the following two documents and analyze which one is better simplified based on the factors of coherence, simplicity, and faithfulness.\\
    Document 1: {\color{blue}{\text{[Simplified document by the vanilla ChatGPT]}}}\\
    Document 2: {\color{blue}{\text{[simplified document by the target method]}}}\\
    In your analysis, please consider how well each document maintains:\\
    - Coherence: The logical flow and organization, ensuring smooth transitions between sentences and paragraphs.\\
    - Simplicity: The level of complexity and difficulty, aiming to make the content more accessible through plain language, shorter sentences, and simpler vocabulary.\\  
    - Faithfulness: How well each document preserves the core meaning, key information, and intended message of the original document without distorting or misrepresenting them.\\
   Please note that you must provide some thoughts and analysis on comparing the two documents, and in the last line, present the improved simplified document.\\
   Follow the output format: [Reasoning content: ... The better-simplified document: (Document 1 or Document 2)]
\end{boxedminipage}
\caption{Example of using prompting template to compare the effectiveness of simplified documents generated using vanilla ChatGPT and simplified documents generated by the target method being tested, where the contents within "[]" are variables.}
\label{fig:Judge}
\end{figure}

\textbf{Baselines.} We compare our methods (based on gpt-3.5-turbo~\footnote{https://platform.openai.com/docs/models/gpt-3-5})  SumDS and ProgDS~\footnote{For samples that exceed the token limit of this interface, we utilize gpt-3.5-turbo-16k to obtain a longer context window.} with the following four baselines.

(1) Keep it Simple (KIS): An unsupervised method for paragraph-level text simplification \citep{laban2021keep}.

(2) BART-SWI: A model fine-tuned on the SWIPE dataset, which pairs documents from English and Simple English Wikipedia for large-scale document-level simplification \citep{laban2023swipe}.

(3) PG$_{\text{Dyn}}$: A plan-guided system where a planner predicts sentence-level operations and provides control tokens to a BART simplification model \citep{cripwell2023document}.

(4) ChatGPT with three prompt templates (P1, P2, and IC): Utilizing gpt-3.5-turbo, three prompt templates are used to sample at a temperature of 0.3. P1 is a basic document-level simplification prompt, P2 emphasizes the distinction between document simplification and summarization, and ChatGPT+IC includes complex-simple document pairs for enhanced in-context learning. The prompts are detailed in Appendix A.

\subsection{Comparison of DS methods} 

\begin{table}[t!]
\small
\centering
\caption{\label{tab:main} Experimental results of document simplification on Wiki-auto. "IC” represents the inclusion of few-shot and chain-of-thought settings. "IT” indicates the final simplified result of the model with the iteration (set to 2). The best results are indicated in bold.}
\begin{tabular}{l|ccccc}\hline
 \textbf{Method} & \textbf{SA} & \textbf{DSA } & \textbf{BAR } & \textbf{FKG $\downarrow$} & \textbf{GPT} \\\hline
 Reference  & 98.81 & 94.14  & -2.93 & 6.55  & 89.63  \\
\cdashline{1-6} 
 KIS  & 32.63 & 29.72  & -3.78 & 8.48  & 44.58  \\
 BART-SWI  & 37.38 & 30.65 & -3.02 & 7.63 & 62.34 \\
 PG$_{\text{Dyn}}$  & 35.61 & 27.83 & -3.83 & 8.86 & 58.42\\
 ChatGPT-P1 & 37.25 & 24.32 & -3.53 & 6.16 & - \\
 ChatGPT-P2 & 37.31  & 22.12 & -3.27 & 6.17 & 56.98 \\
 ChatGPT+IC & 38.03 & 23.34 & -3.12 & 7.51 & 63.28 \\
\cdashline{1-6} 
 SumDS & 38.73 & 27.58 & -2.98 & 7.88 & 67.76\\
 SumDS+IC & 39.72 & 29.83 & -3.11 & 6.79 & 72.53 \\
 ProgDS & 42.33  & 33.79 & -2.73 & 6.35 &  69.53\\
 ProgDS+IC & 44.68  & 35.26 & -2.78 & 6.79 & 77.92 \\
 ProgDS+IT & \textbf{45.83}  & \textbf{38.46} & \textbf{-2.53} & \textbf{5.96} & \textbf{79.43} \\
\hline
\end{tabular}

\end{table}

\begin{table}[h!]
\small
\centering
\caption{\label{tab:main2} Experimental results of document simplification on Newsela. The best results are indicated in bold.}
\begin{tabular}{l|ccccc}\hline
 \textbf{Method} & \textbf{SA} & \textbf{DSA } & \textbf{BAR } & \textbf{FKG $\downarrow$} & \textbf{GPT} \\\hline
  \multicolumn{6}{c}{Newsela-A} \\\hline
Reference  & 65.82 & 48.05 & -2.57 & 6.61  &  87.63\\
\cdashline{1-6} 
 KIS  & 33.26 & 26.58  & -2.92 & 9.32 & 42.77  \\
 BART-SWI  & 30.23 & 23.78 & -3.16 & 8.58 & 45.68 \\
 PG$_{\text{Dyn}}$   & 43.33 & 36.62 & -3.03 & 6.74 & 62.35\\
 ChatGPT-P1 & 34.23 & 23.43 & -3.26 & 6.98 & - \\
 ChatGPT-P2 & 35.98 & 24.20 & -3.31 & 9.54 & 65.48 \\
 ChatGPT+IC & 35.26 & 24.85 & -2.80 & 11.23 & 71.42 \\
\cdashline{1-6} 
 SumDS & 34.10 & 26.23 & -2.92 & 6.32 & 65.35\\
 SumDS+IC & 35.63 & 25.27 & -2.48 & 6.75 & 69.73 \\
 ProgDS & 44.35  & 38.53 & -2.87 & 5.78 & 72.03 \\
 ProgDS+IC & 46.53  & 39.28 & -2.38 & 5.89 & 76.59 \\
 ProgDS+IT & \textbf{46.89} & \textbf{40.36} & \textbf{-2.28} & \textbf{5.62} & \textbf{82.53} \\
\hline
 \multicolumn{6}{c}{Newsela-B} \\\hline
 Reference  & 67.25 & 49.83  & -2.13 & 6.87  & 88.34  \\
\cdashline{1-6} 
 ChatGPT-P1 & 31.35 & 21.75 & -3.24 & 7.66 & - \\
 ChatGPT-P2 & 32.18 & 26.64 & -2.65 & 9.78 & 73.93 \\
 ChatGPT+IC & 32.48 & 25.04 & -2.81 & 11.34 & 75.85 \\
\cdashline{1-6} 
 SumDS & 35.27 & 29.58 & -3.12 & 6.68 & 68.35 \\
 SumDS+IC & 37.53 & 32.67 & -2.96 & 7.68 & 69.56 \\
 ProgDS & 40.26  & 36.62 & -3.42 & 7.15 & 72.69 \\
 ProgDS+IC & 42.48  & 37.56 & -2.68 & 6.33 & 75.27 \\
 ProgDS+IT & \textbf{43.35} & \textbf{37.68} & \textbf{-2.48} & \textbf{6.83} & \textbf{78.96} \\
\hline
\end{tabular}
\end{table}

\begin{figure}[t!]   
\small
    \centering
    \includegraphics[width=1.0\linewidth]{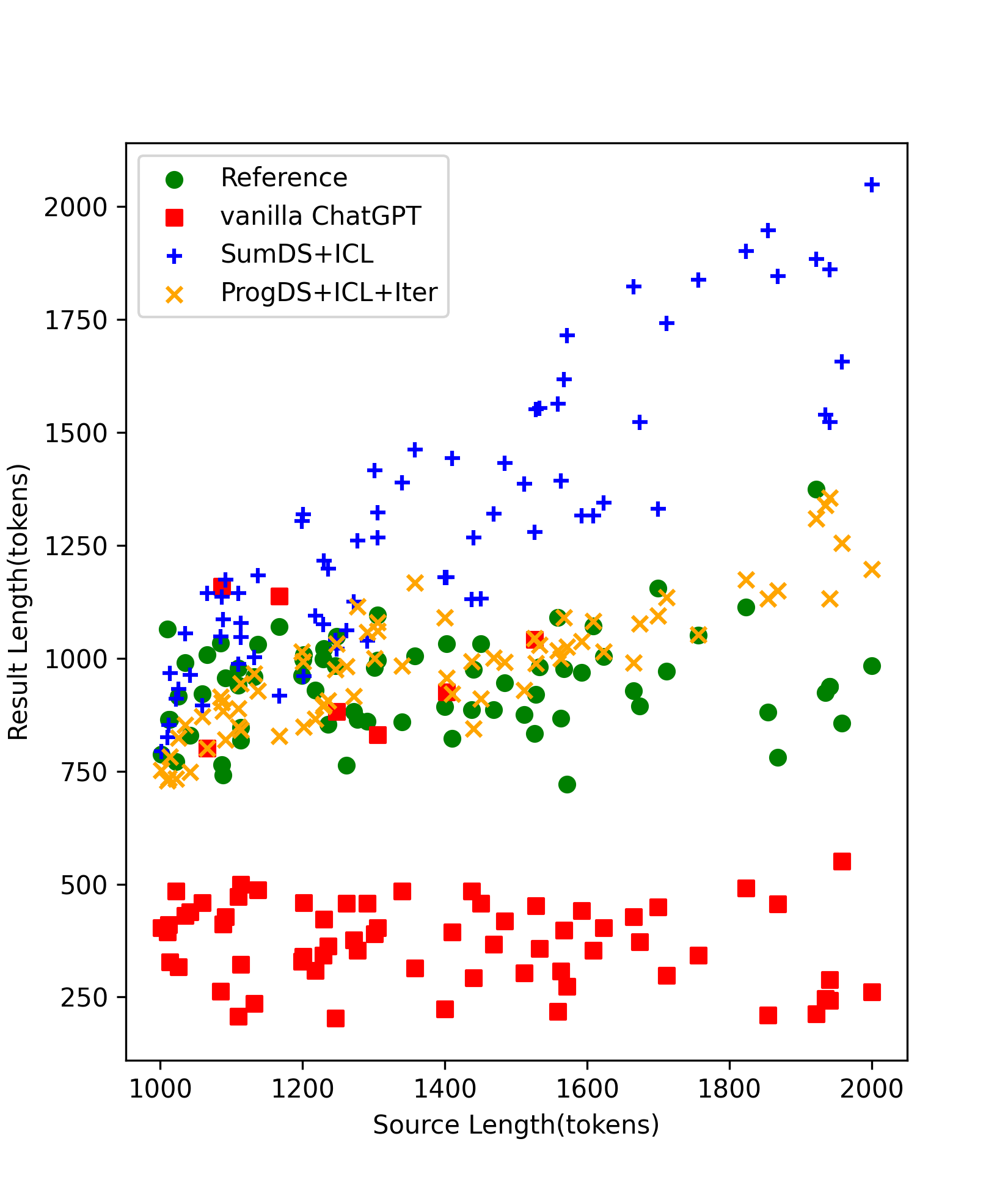}
    \caption{This scatter plot illustrates the comparison of token counts between documents generated by several different methods and the reference document.}
    \label{fig:scatter}
\end{figure}

Experimental results across three datasets in Table \ref{tab:main} and Tabel \ref{tab:main2} show that SumDS and ProgDS outperform traditional small-scale models and direct use of ChatGPT in nearly all evaluation metrics. For Newsela datasets, average scores of multiple reference documents based on metrics like SARI and D-SARI were used. As document length increases, the performance of SumDS and ProgDS becomes more pronounced, while small-scale models struggle with longer documents and are excluded from Newsela-B results.

Both SumDS and ProgDS achieve high BART-Score, SARI, and FKGL scores, indicating effective removal and replacement of complex content while maintaining source faithfulness. ChatGPT's performance with prompts (P1, P2, and IC) is poor for long documents, typically producing outputs under 500 tokens, far shorter than the reference documents. This issue persists even with in-context learning, suggesting a misunderstanding of document simplification as summarization.

ProgDS outperforms SumDS, particularly in higher editing rates and overall document coherence and readability. ProgDS's output lengths are closer to reference documents due to the effective removal of redundant paragraphs, especially in longer source documents, aligning more closely with human experts' approach.

Additionally, few-shot learning and chain-of-thought methods significantly improve automatic evaluation metrics and ChatGPT success rates, demonstrating that contextual learning and reasoning examples enhance task understanding and output quality.

Compared to traditional methods of pre-training then fine-tuning, as well as directly prompting LLMs, our proposed framework has the following three advantages:

(1) LLMs are trained on massive amounts of textual data, which gives them a superior understanding of copy-editing tasks. Furthermore, as LLMs undergo instruction-tuning with human preferences, clever prompt templates can be designed to accomplish our tasks, eliminating the high data and computational requirements of traditional methods.

(2) ProgDS follows the simplification methods of human experts, integrating discourse-level, topic-level, lexical-level simplification based on the principles of content and hierarchical division. This alleviates the limitations of LLMs unable to simplify long documents through direct prompting.

(3) ProgDS achieves more comprehensive simplification results, with a higher modification rate compared to traditional methods, rather than excessive retention. Additionally, we can determine the level of simplification iteratively, offering a higher quality and more flexible solution than traditional methods.

\textbf{Human evaluation:} To obtain more accurate and user-friendly evaluation results, we also need to conduct manual evaluations on the simplified documents beyond various automated evaluation metrics. We have hired three graduate students with good English proficiency to complete the evaluation. We selected the SumDS method and the ProgDS method as the primary test subjects and used the PG\textsubscript{Dyn} method and reference documents as comparisons (to strike a balance between simplicity and faithfulness, we chose the second version from four simplified reference documents). 

We randomly select 50 document pairs from Newsela-A and fed the source documents into both the PG\textsubscript{Dyn} model and the ProgDS method to obtain test samples. We do not use the Newsela-B dataset because the PG\textsubscript{Dyn} model cannot handle the document length in this dataset. During the manual evaluation, we use a five-point Likert scale to measure the quality of the simplified documents in three aspects: Coherence, Simplicity, and Faithfulness. Here's a shorter explanation of each evaluation aspect:

\begin{itemize}
\setlength{\itemsep}{1pt}
    \item Coherence: Evaluates the logical flow and organization of the simplified document, ensuring smooth transitions and interconnectedness between sentences and paragraphs.
    \item Simplicity: Assesses the level of complexity and difficulty in the simplified document, aiming to make the content more accessible through plain language, shorter sentences, and simpler vocabulary.
    \item Faithfulness: Measures how well the simplified document preserves the core meaning, key information, and intended message of the original document without distorting or misrepresenting them.
\end{itemize}

\begin{table}[t!]
\small
\centering
\resizebox{\linewidth}{!}
{
\begin{tabular}{l|cccc}\hline
\textbf{Methods} & \textbf{Coherence} & \textbf{Simplicity} & \textbf{Faithfulness} & \textbf{Average} \\\hline
PG\textsubscript{Dyn} & 2.97 & 3.89 & 2.85 & 3.24 \\
SumDS+ITE & 3.32 & 3.28 & \textbf{4.23} & 3.58 \\
ProgDS+ITE & 4.55 & 4.02 & 3.75 & 4.11 \\
Reference & \textbf{4.63} & \textbf{4.25} & 3.97 & \textbf{4.28} \\\hline
\end{tabular}
}
\caption{\label{tab:human} Results of human evaluation on 50 documents randomly selected from Newsela-A. The best results are shown in bold. "Average" represents the average scores of the three aspects. }
\end{table}

The human evaluation results are shown in Table \ref{tab:human}. It can be observed that SumDS and ProgDS (with the support of IC and IT), align closely with human evaluators' preferences in almost three aspects, and the overall score can even surpass the reference document. Among them, the coherence score of the ProgDS method is higher, indicating that the approach of segmenting long documents based on hierarchy is superior to the approach based on content in terms of overall coherence. The SumDS method achieves a higher score in the faithfulness metric, which could be attributed to its approach of not deleting entire paragraphs, thus adhering more closely to the original content. 

The simplified articles edited using LLMs prompting framework have achieved a reading experience that is nearly on par with or even surpasses articles edited by human experts. The evaluation indicates that although LLMs like ChatGPT may lack extensive training data and instructions on DS tasks, their performance on such complex tasks can be improved through a fixed framework.

We also add manual error analysis to the ablation study to see when the approach fails without a particular step will also be helpful. We randomly select 50 document pairs from Newsela-A. 

We conduct more detailed manual evaluations as shown in Table \ref{tab:eval_ablation} to analyze the impact of each simplification stage on readability. 

\begin{table}[h]
\small
\centering
\resizebox{\linewidth}{!}
{
\begin{tabular}{l|cccc}\hline
 & \textbf{Coherence} & \textbf{Simplicity} & \textbf{Faithfulness} & \textbf{Avgerage}  \\\hline
ProgDS  & \textbf{4.32} & \textbf{3.97} & 3.54 & \textbf{3.94} \\
-w/o Discourse  & 3.66 & 3.28 & \textbf{3.75} & 3.56 \\
-w/o Topic & 3.87  & 3.42 & 3.48  & 3.59 \\
-w/o Lexical & 4.13 & 3.18 & 3.32 & 3.54 \\\hline
\end{tabular}
}
\caption{Results of human evaluation on 50 documents randomly selected from Newsela-A.}

\label{tab:eval_ablation}
\end{table}

We can summarize the following conclusions: 

(1) The discourse-level simplification stage improves coherence and simplicity scores by reordering paragraphs and generating subtitles. However, excessive paragraph removal may affect faithfulness.

(2) The topic-level simplification stage enhances simplicity by simplifying structures and sentences while maintaining context. It also addresses issues like anaphora choice.

(3) The lexical-level simplification stage improves simplicity by replacing complex expressions with simpler ones, significantly enhancing readability.

\subsection{Ablation Study} 

 To validate our proposed framework, we conducted an ablation experiment using 300 samples from the Newsela-A dataset, chosen for its shorter document length, which was suitable for verifying the necessity of different framework components. 

The results, shown in Table~\ref{tab:ablation}, indicate that removing the \textit{summary generation} stage in the SumDS method decreases performance, suggesting that the summary document guides paragraph simplification. This guidance ensures inter-paragraph consistency and helps the model generate essential sentence and lexical edits, enhancing the faithfulness of the simplified document.

When the discourse-level simplification stage of ProgDS is removed, there are minimal changes in performance, with a slight improvement in the SARI and D-SARI metrics. This stage seems to improve coherence and readability at a high level, which enhances reader experience but is not well-captured by automated metrics. In contrast, removing the topic-level and lexical-level simplification stages significantly reduces performance, particularly in lexical-level simplification, which impacts the automated evaluation metrics more substantially. Thus, all three stages are essential from a practical application perspective.

\begin{table}
\small
\centering
\begin{tabular}{l|cccc}\hline
 & SARI  & D-SARI  & BART-S & FKGL  \\\hline
SumDS  & \textbf{35.27} & \textbf{29.58} & \textbf{-3.12} & \textbf{6.68}\\
-w/o Summary & 33.82  & 26.76 & -3.85 & 7.34 \\
\hline
ProgDS  & 40.26  & 36.62 & -3.42 & \textbf{7.15} \\
-w/o Discourse  & \textbf{41.57} & \textbf{36.84}  & -3.73 & 8.32 \\
-w/o Topic & 37.26  & 32.75 & -3.59  & 7.68 \\
-w/o Lexical & 23.63 & 14.51 & \textbf{-2.98} & 8.79 \\\hline
\end{tabular}
\caption{Ablation study on the SumDS and ProgDS methods using the Newsela-A dataset. "-w/o" indicates the method without the specific stage. }
\label{tab:ablation}
\end{table}

\begin{figure}[h!]
\small
    \centering
    \caption{Relationship of time spent or number of model calls when varying the length of the document.}
    \includegraphics[width=1.0\linewidth]{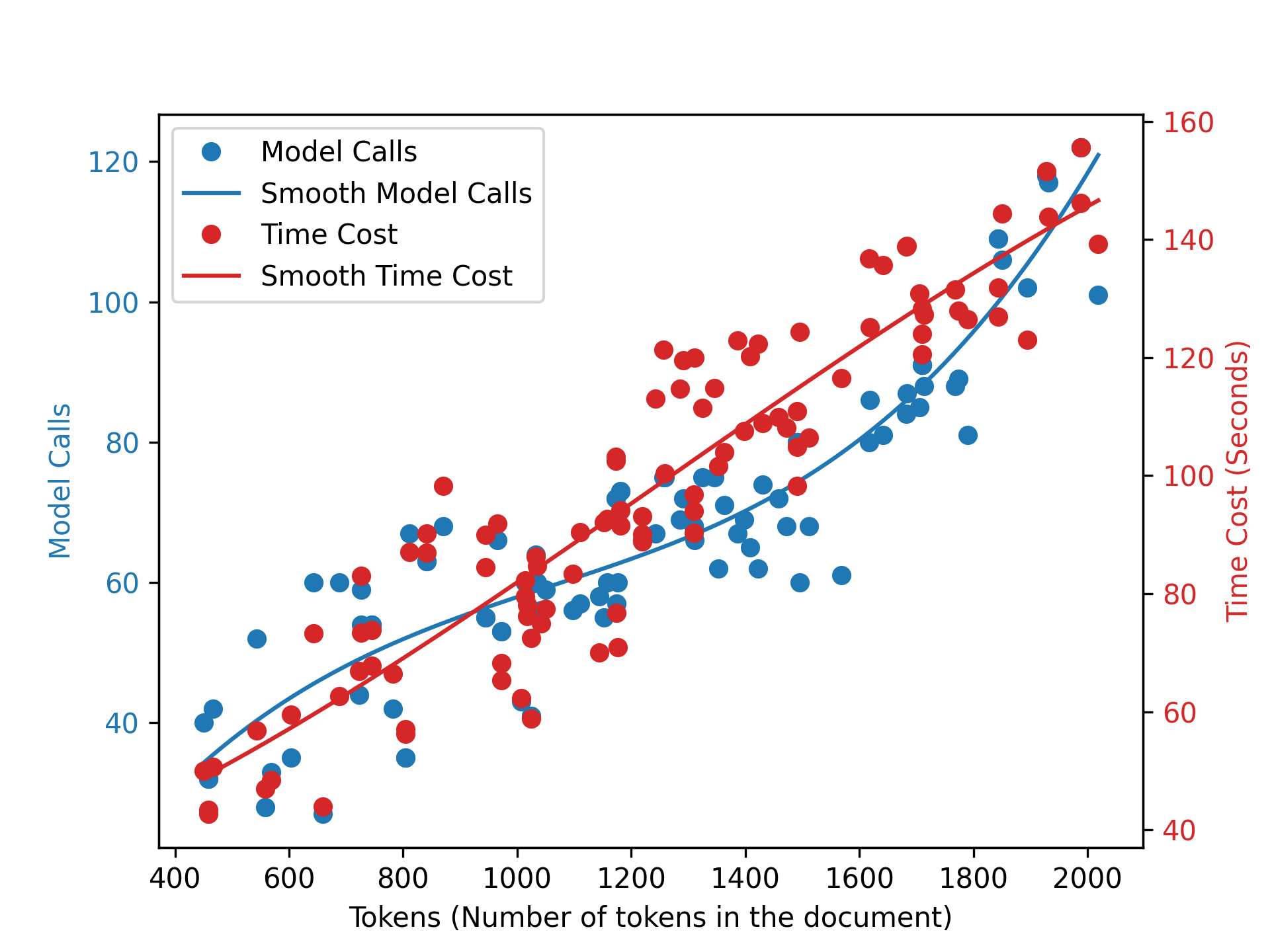}
    \label{fig:relation}
\end{figure}

\textbf{Complexity of ProgDS: } How does the computational complexity of ProgDS scale with the length of the input document? We analyze ProgDS in terms of time spent and the number of model calls when varying the length of the document. The results are shown in Figure \ref{fig:relation}. As the length of the text increases, the number of model calls and the time spent also increase. One limitation is that when the document length extends considerably, leading to a gradual increase in the number of calls to LLMs.

\section{Conclusions}

We introduce one novel LLM-based long document-level text simplification method ProgDS, which can generate simplified versions of complex long documents to improve document readability while striving to meet human preferences. Specifically, due to the lack of adaptability of LLMs in the task of simplifying long documents, we decompose the task into multiple subtasks, allowing LLMs to progressively simplify from shallow to deep. Additionally, We mitigate the hallucinations generated by LLMs using the over-generate-then-filter approach and enhance document simplification through iterative simplification. Experimental results demonstrate that our proposed frameworks outperform previous DS methods and have a wider acceptance range. In the future, we will attempt to apply our method to the Multilingual DS task. We will also explore the application of this method to other long document text-processing tasks.





\bmhead{Acknowledgements}

This research is partially supported by the National Natural Science Foundation of China (62076217), and the National Language Commission of China (ZDI145-71). 

\bmhead{Data availability} 

The data used in this paper are all from public datasets.

\bmhead{Author Contributions} Dengzhao Fang: Methodology. Jipeng Qiang: Idea, investigation, Writing - original draft. Yi Zhu and Yunhao Yuan: Writing - review \& editing. Wei Li and Yan Liu: Supervision, Resources.

\section*{Declarations}

\textbf{Competing interests} The authors declare that they have no known competing financial interests or personal relationships that could have appeared to influence the work reported in this paper.

\textbf{Ethical Standard} The article was submitted with the consent of all the authors to participate.

\begin{appendices}

\section{The prompting template of ChatGPT}

\textbf{ChatGPT-P1.} The prompting template of ChatGPT-P1 is shown in Figure~\ref{fig:prompt_base}.

\begin{figure}[h!]
\centering
\begin{boxedminipage}{\columnwidth}
\footnotesize
You are a professional simplified text writer, I need you to simplify the language and structure of the raw text to make it more accessible to pupils. \\
Replace complex words or phrases or technical terms with simpler, more familiar words or terms, use more and shorter clauses, and reorganize clauses to make them easier to read. \\
Raw text: \\
{\color{blue}{\text{[Raw text]}}}\\
Simplified text:
\end{boxedminipage}
\caption{The prompting template of \textit{P1} for DS task. Basic document-level simplification prompts without contextual examples.}
\label{fig:prompt_base}
\end{figure}

\textbf{ChatGPT-P2.} The prompting template of ChatGPT-P2 is shown in Figure~\ref{fig:prompt_force}.

\begin{figure}[h!]
\centering
\begin{boxedminipage}{\columnwidth}
\footnotesize
As a text simplification writer, your task is to simplify the given text content: restate the original text in simpler and easier-to-understand language without changing its meaning as much as possible. \\
You can change paragraph or sentence structure, remove some redundant information, and replace complex and uncommon expressions with simple and common ones. \\
It should be noted that the task of text simplification is completely different from the task of text summarization, so you need to provide a simplified parallel version based on the original text, rather than just providing a brief summary. \\
Raw text: \\
{\color{blue}{\text{[Raw text]}}}\\
Simplified text:
\end{boxedminipage}
\caption{The prompting template of \textit{P2} by emphasizing the difference between document simplification task and summary task for DS task. On the basis of the basic prompt as shown in Figure\ref{fig:prompt_base}, emphasize the difference between the document simplification task and summary task and guide the model to generate a parallel simplified version of the original text without contextual examples.}
\label{fig:prompt_force}
\end{figure}

\textbf{ChatGPT+IC.} The prompting template of ChatGPT+IC is shown in Figure~\ref{fig:prompt_oneshot}.

\begin{figure}[h!]
\centering
\begin{boxedminipage}{\columnwidth}
\footnotesize
{\color{blue}{\text{[(A Complex Document - Simple Document Example)]}}}\\
Now please study the example above and simplify the document below. Please note that document simplification is not a document summary. You cannot shorten the original text to a very small length. \\
The operations you need mainly include paragraph order reconstruction, redundant information removal, sentence structure simplification, and replacing complex words or phrases with simple expressions. In addition, simplified documents require subheadings starting with \#\# to improve readability. \\
Raw text: \\
{\color{blue}{\text{[Raw text]}}}\\
Simplified text:
\end{boxedminipage}
\caption{The prompting template of \textit{ChatGPT+IC} with in-context learning for DS task. Based on the two prompting templates as shown in Figure\ref{fig:prompt_base} and Figure\ref{fig:prompt_force}. Additionally, a pair of complete complex-simple documents are added to drive the in-context learning ability of the large language model.}
\label{fig:prompt_oneshot}
\end{figure}

\section{Case Study}

\begin{figure*}[ht!]   
    \centering
    \includegraphics[width=\linewidth]{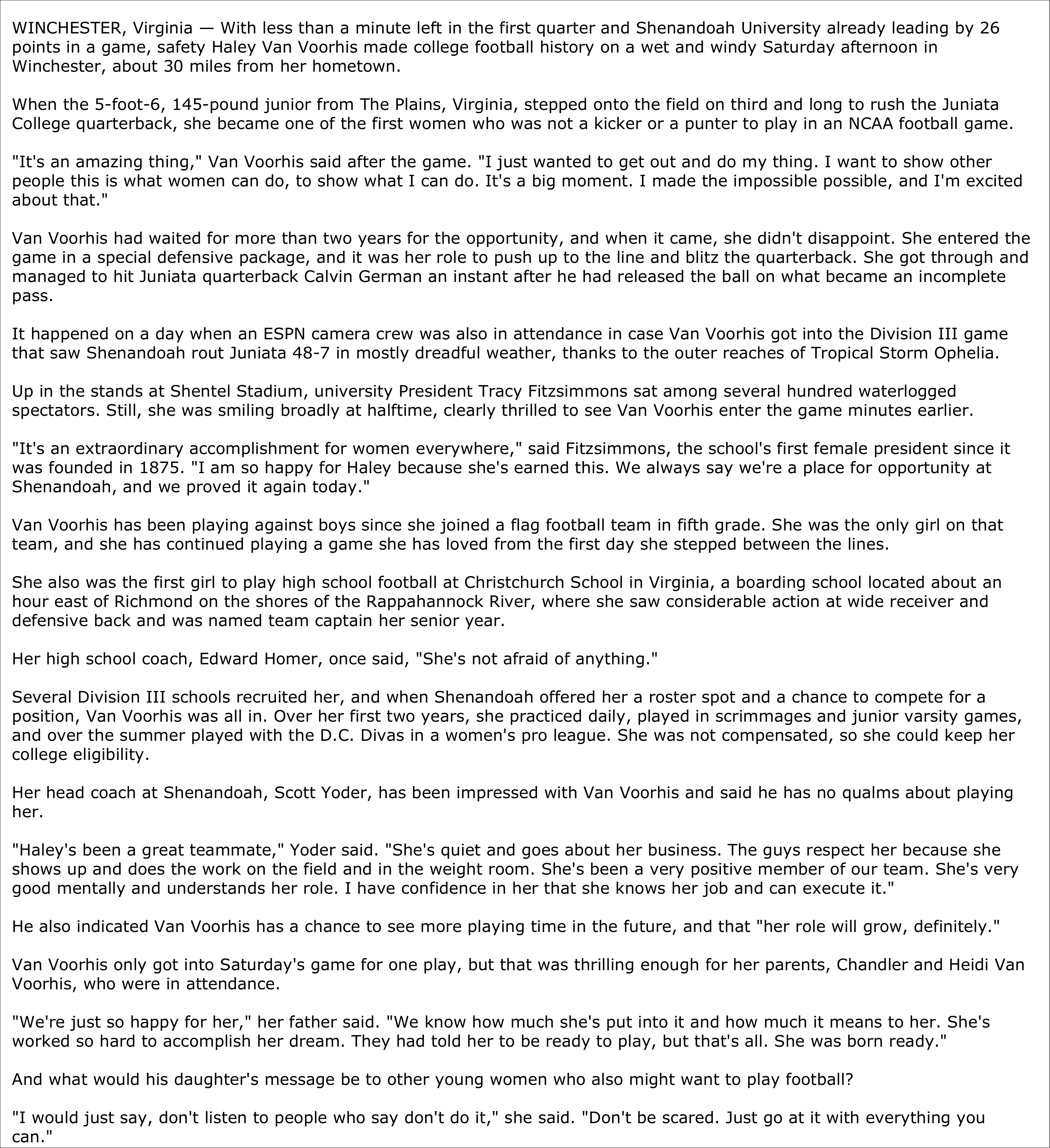}
    \caption{The source document extracted from the Newsela official website.}
    \label{fig:Raw_doc}
\end{figure*}

\begin{figure*}[h]   
    \centering
    \includegraphics[width=\linewidth]{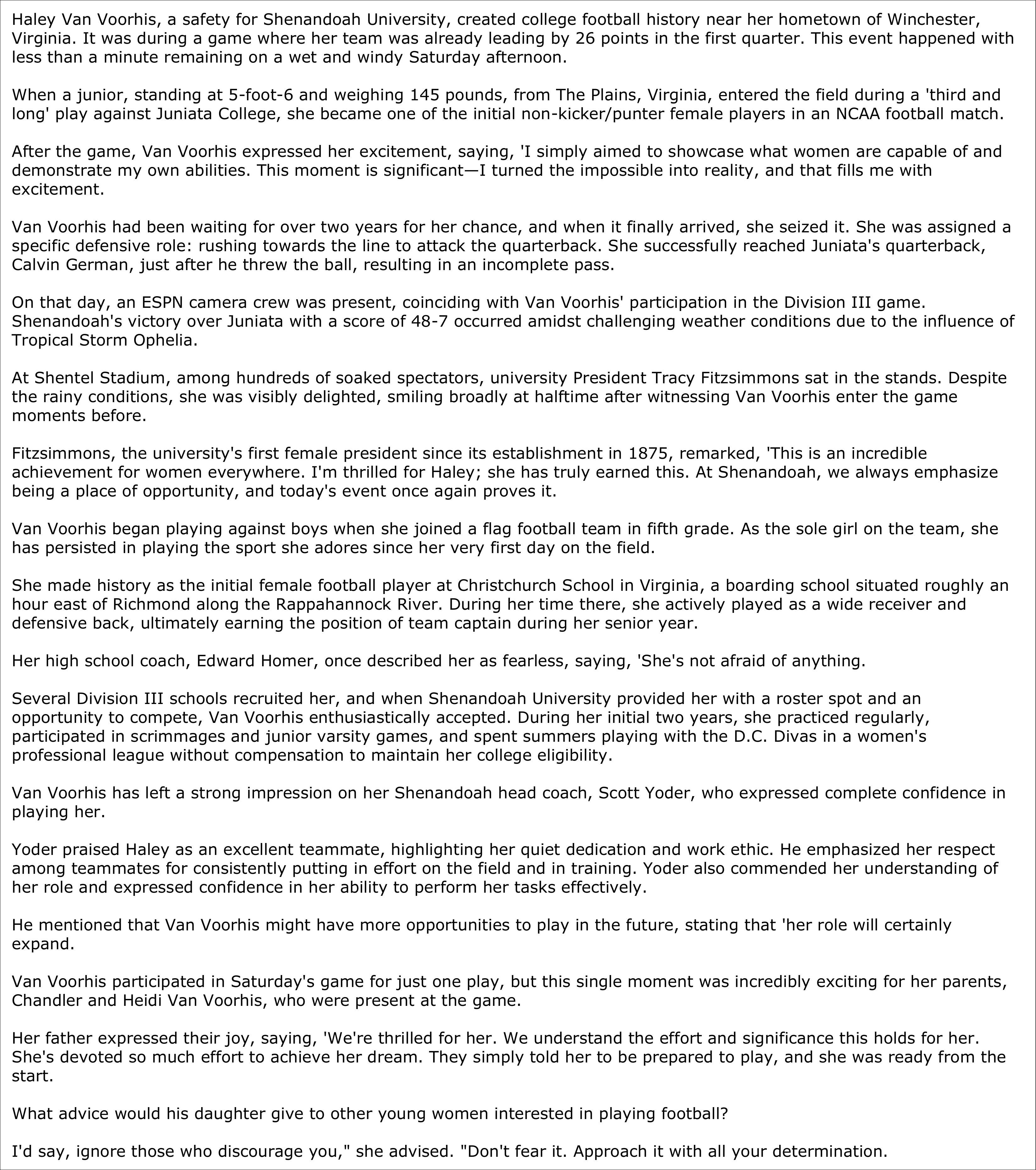}
    \caption{The simplified document generated by the SumDS framework under zero-shot.}
    \label{fig:SumDS_doc}
\end{figure*}

\begin{figure*}[ht]   
    \centering
    \includegraphics[width=\linewidth]{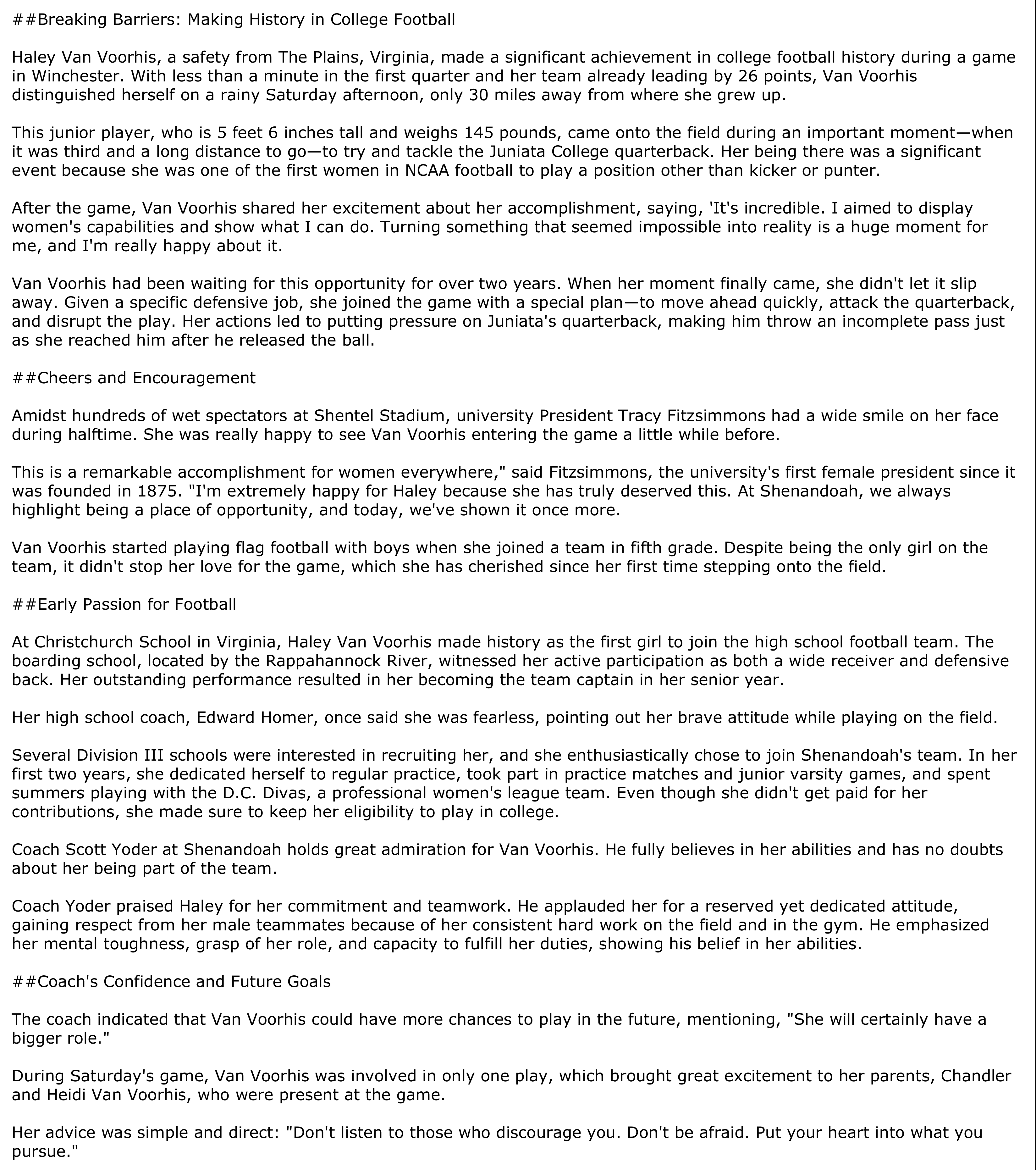}
    \caption{The simplified document generated by the ProgDS framework under zero-shot.}
    \label{fig:ProgDS_doc}
\end{figure*}

\begin{figure*}[h]   
    \centering
    \includegraphics[width=\linewidth]{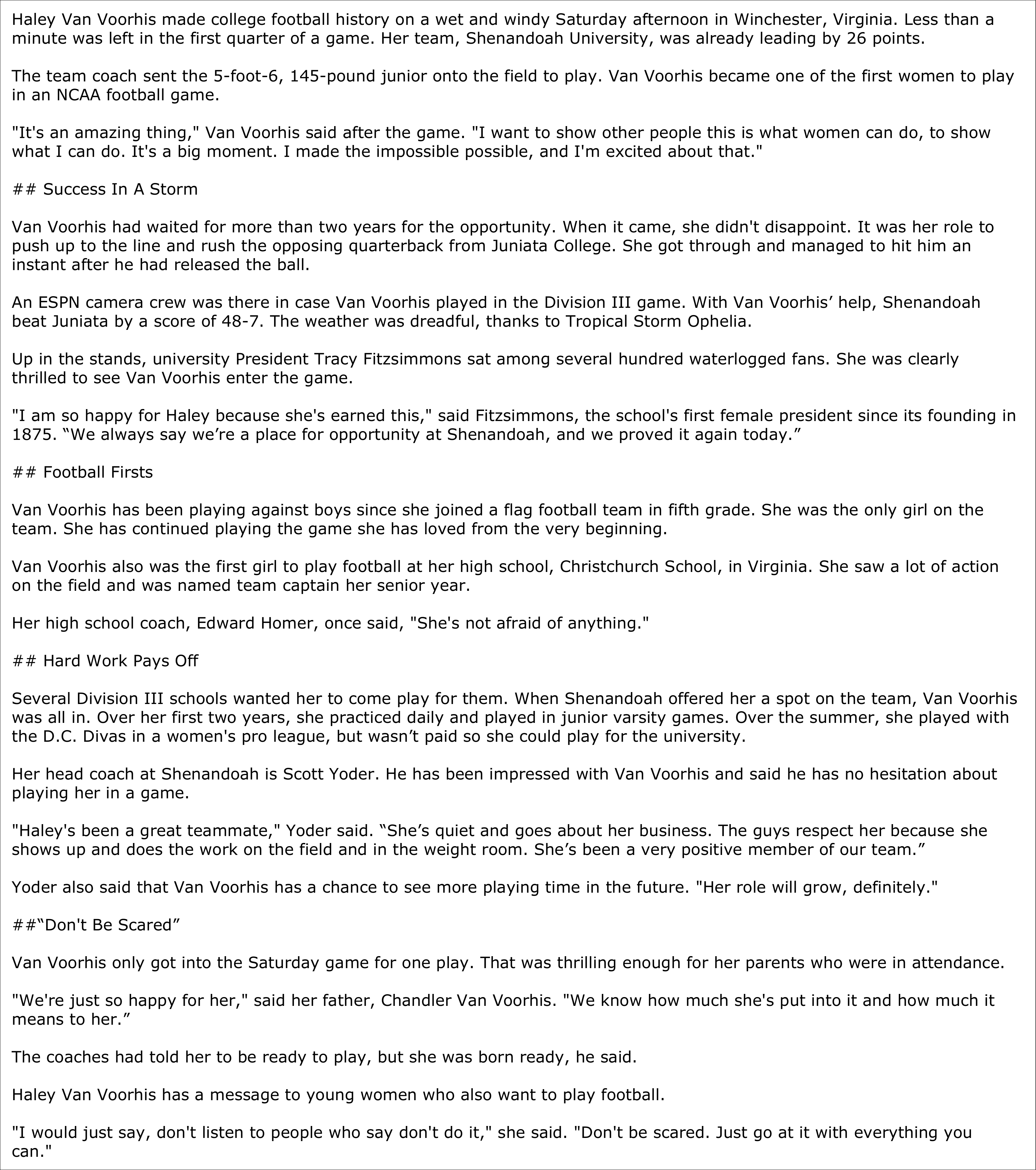}
    \caption{The simplified document extracted from the Newsela official website.}
    \label{fig:Ref_doc}
\end{figure*}

Here we present examples of document simplification using the SumDS and ProgDS methods, located in Figure \ref{fig:SumDS_doc} and Figure 
\ref{fig:ProgDS_doc}, respectively. The source document is shown in Figure \ref{fig:Raw_doc} and the reference document is shown in Figure \ref{fig:Ref_doc}.

\vspace{4cm}

\vspace{4cm}

\vspace{4cm}


\end{appendices}

\clearpage
\bibliography{sn-article}

\end{document}